%% file: main.tex
\documentclass[pdflatex,sn-apa]{sn-jnl}

\usepackage{graphicx}
\usepackage{amsmath,amssymb,amsfonts}
\usepackage{amsthm}
\usepackage{booktabs}
\usepackage{enumitem}

\theoremstyle{thmstyleone}
\newtheorem{proposition}{Proposition}
\theoremstyle{thmstylethree}

\begin{document}

\title[Affective Sovereignty as an Epistemic Consequence]{Who Determines the Meaning of an Emotion? Affective Sovereignty as an Epistemic Consequence of Measurement Limits}

\author[1]{\fnm{Keito} \sur{Inoshita}}\email{inosita.2865@gmail.com}

\affil[1]{\orgdiv{Faculty of Business and Commerce}, \orgname{Kansai University}, \orgaddress{\street{3-3-35 Yamate-cho}, \city{Suita}, \postcode{564-8680}, \state{Osaka}, \country{Japan}}}

\abstract{\input{abstract_en}}


\maketitle

\input{body_en}

\backmatter

\bmhead{Declarations}

\paragraph{Funding}
This work was supported by JST SPRING, Japan, Grant Number JPMJSP2150.

\paragraph{Competing interests}
The author declares no competing interests.

\paragraph{Data availability}
No new datasets were generated or analysed in this study.

\bibliography{refs}

\end{document}

%% file: abstract_en.tex
Emotion-sensing AI is rapidly becoming embedded in vehicles, home appliances, dialogue agents, and social infrastructure, giving rise to a sphere in which emotion is no longer confined to individual experience but is instead observed and computed at a societal scale, a domain we term the Affectosphere. Yet a central normative question in this domain has remained underexplored: who has the final authority to determine the meaning of one's own emotion? This study addresses the question from the epistemological side of measurement's structural limits. We define a meaning distribution as the distribution of labels assigned by annotators drawn from a population under a fixed annotation protocol, and decompose its uncertainty into reducible and irreducible components. We then demonstrate that, while emotion AI can assign high-confidence point labels and discriminate real differences at an aggregate level, the irreducible component of the meaning distribution for individual instances cannot be estimated with adequate coverage under realistic annotator counts, a systematic divergence we term the epistemic gap. The key finding is that high device confidence does not constitute evidence that irrecoverable meaning has been recovered. From this epistemic gap, together with an explicitly stated normative premise, namely that the output of a system which cannot recover a quantity in principle must not be treated as its authoritative determination, we derive the norm that the final interpretive authority over the meaning of one's emotion is procedurally reserved for the experiencing subject, the norm of affective sovereignty. These results suggest that the design, evaluation, and regulation of emotion AI should place explicit allocation of interpretive authority, rather than accuracy maximisation, at their core.

%% file: body_en.tex
\section{Introduction}\label{sec:intro}

Artificial intelligence (AI) that reads and infers emotion has, over the past several years, been repositioned from a laboratory technology to a foundational component of society. The attempt to make emotion an object of computation, systematised as the field of affective computing~\citep{picard1997affective}, has matured into systems that estimate emotional states from diverse signals: text, facial expressions, speech, physiological responses, and the rhythm of conversation. Such systems have been incorporated into call-centre support, in-vehicle driver monitoring, social-media analytics, and dialogue agents for education and healthcare, creating a situation in which emotion is measured in virtually every domain of daily life~\citep{mcstay2018emotional,andalibi2020human}. This ubiquity means that emotion is no longer a private experience confined to the individual interior; it is being woven into the fabric of social structures and becoming observable and computable. We term this domain the Affectosphere. The metaphor extends the idea of emotion as an atmospheric quality of social space~\citep{anderson2009atmospheres} and is the affective counterpart of the infosphere in which information becomes pervasive and reshapes human reality~\citep{floridi2014fourth}. Whereas twentieth-century emotion research consistently located emotion ``within the individual,'' the rise of emotion AI is quietly, yet fundamentally, rewriting that premise.

The advent of the Affectosphere simultaneously expands convenience and inaugurates new normative problems. The more widely technology for measuring emotion spreads, the more the emotion labels produced by measurement circulate in advance of the individual's own self-understanding, and the more emotion itself becomes a resource that platforms measure and commodify~\citep{zuboff2019surveillance}. In an environment where AI continuously and unilaterally interprets ``you are angry'' or ``you seem sad,'' the right to determine the meaning of one's own emotion can be silently eroded, not through explicit violation, but as a design default. This is the emotional analogue of the challenges confronted by informational self-determination, the claim of individuals to determine for themselves how information about them is used~\citep{westin1967privacy}, and by respect for autonomy in medicine~\citep{beauchamp2019principles}, and it constitutes the central normative problem of an era in which emotion-processing technologies become social infrastructure. Prior work has responded to this problem largely with defensive frameworks: proposing new rights against the reading of mental states~\citep{ienca2017newrights}, and articulating the ethics of emotion recognition together with critical perspectives on affective AI from data to society~\citep{mohammad2022ethics,inoshita2026emotion}. A recent proposal has even named this right affective sovereignty and built a computational framework to enforce it~\citep{kim2026affective}. Yet that proposal grounds the right ethically and engineers its enforcement; it does not establish why the authority cannot, even in principle, be discharged by an ever-better measuring system. The core of the question thus remains open on its decisive side, the epistemic one: who holds the final interpretive authority over the meaning of emotion, and why can measurement not settle it?

The difficulty in addressing this question directly is that it cannot remain a normative claim alone; it must engage the epistemology of measurement. Emotion AI can output high-confidence point labels and scores that appear well-calibrated for each instance, and at the aggregate level of a population such outputs can meaningfully discriminate real differences. However, whether the ``true meaning'' of an emotion for an individual instance can be recovered from a finite annotation is an entirely separate question. Label assignment inherently varies across annotators, and a growing body of work shows that this variation is not mere noise to be averaged away but in part a genuine signal~\citep{aroyo2015truth,pavlick2019inherent,plank2022variation,uma2021disagreement}; that is, it contains a component that does not vanish as annotator count increases, the aleatoric counterpart of irreducible uncertainty~\citep{hullermeier2021aleatoric}. If the relationship between device confidence and the recoverability of individual meaning is conflated when constructing a normative argument, the discussion of interpretive authority either collapses into speculation that ignores measurement limits, or inadvertently implies that improving measurement precision can substitute for interpretive authority.

In this study, we formalise the problem of interpretive authority over the meaning of emotion from the epistemological side of measurement's structural limits, and derive the locus of that authority as a consequence. The line of argument is as follows. We first set the stage of the Affectosphere, the domain in which emotion is observed and computed like the atmosphere of society, and pose the locus of interpretive authority, namely who ultimately determines the meaning of an emotion, as the central problem. We then operationalise the meaning of emotion as the distribution of labels assigned by annotators under a fixed protocol, and decompose its uncertainty into reducible and irreducible components. We show that the irreducible component cannot be recovered with adequate coverage from finite observations, drawing on the ill-posedness of estimating a distribution when the number of annotators is small relative to the number of labels, and on Jensen's inequality for an unbiased estimator. Finally, by joining the premise that the determination of a quantity must not be delegated to a system that cannot recover it with the asymmetry of access to the context that constitutes the meaning of an emotion, we derive that interpretive authority belongs to the experiencing subject. Through this derivation, this study positions affective sovereignty not as a preference for autonomy or an ethical demand, but as a conclusion from an explicit normative premise together with the structural limits of measurement. The main contributions of this study are summarised as follows.

\begin{enumerate}[label=\roman*),leftmargin=2.2em]
\item A meaning distribution is defined as the distribution of labels assigned by annotators drawn from a population under a fixed annotation protocol, and a framework is provided that decomposes its uncertainty into a reducible component, which vanishes as the number of annotators grows, and an irreducible component, which persists. This decomposition maps the ambiguity of emotion labels onto the distinction between epistemic and aleatoric uncertainty in machine learning.
\item The epistemic gap, the systematic divergence between emotion AI confidence and the recoverability of individual-instance meaning, is formalised as a proposition. The proposition is shown to rest on a statistical fact that is independent of any particular dataset, and empirical results reported in the related literature are connected as corroboration.
\item Affective sovereignty, the norm that the final interpretive authority over the meaning of one's emotion is procedurally reserved for the experiencing subject, is derived from the epistemic gap. The derivation does not move from fact to ought without mediation, but makes explicit the normative premise of interpretive non-delegation before drawing the conclusion.
\end{enumerate}

The rest of this paper is organised as follows. Section~\ref{sec:related} reviews related work. Section~\ref{sec:gap} introduces the Affectosphere, defines the problem, and presents an overview of the epistemic gap. Section~\ref{sec:formal} formalises irreducibility. Section~\ref{sec:demo} provides empirical grounding. Section~\ref{sec:objections} considers objections and replies. Section~\ref{sec:discussion} discusses implications and limitations, and Section~\ref{sec:conclusion} concludes.

\section{Related Work}\label{sec:related}

\subsection{Neurorights, Affective Privacy, and the Regulation of Emotion AI}

The attempt to make emotion an object of computation has been systematised as the field of affective computing~\citep{picard1997affective}. Alongside progress in that field, new rights frameworks for mental states have been proposed. \citet{ienca2017newrights} advanced four new human rights, cognitive liberty, mental privacy, mental integrity, and psychological continuity, envisioning a normative barrier against technological intervention in the mind. \citet{farahany2023battle} offered a systematic account of cognitive liberty as the right to protect freedom of thought. From the philosophy of law, \citet{bublitz2014crimes} raised a human right to mental self-determination and framed intervention in the mind as a rights violation. These contributions are important for having established the need for rights in an era in which mental states can be read by technology.

Research focused specifically on emotion has developed primarily through the lens of privacy and surveillance. \citet{mcstay2018emotional} discussed emotion AI as the rise of empathic media, pointing to surveillance risks posed by the measurement and use of emotion and the need for soft-law regulation. At the regulatory level, institutional responses to emotion recognition have appeared: the European Union's AI Act includes provisions restricting the use of emotion recognition in workplace and educational contexts, imposing legal constraints on the measurement of emotion itself~\citep{euaiact2024}. These contributions provide a defensive framework of emotion-data protection and use restriction.

The ethical implications of emotion AI have also accumulated. \citet{mohammad2022ethics} systematically organised the ethics of automatic emotion recognition, and \citet{stark2021emotion} structured the ethical issues in AI systems that handle emotion. \citet{mcstay2020emotional} discussed privacy concerns around emotion AI monitoring emotional life as soft biometrics, and \citet{andalibi2020human} empirically clarified how users perceive and experience risks from emotion recognition on social media. However, these frameworks ask who protects emotion data and where emotion AI may be used; they do not directly address the locus of interpretive authority, who has the final say over the meaning of emotion. Moreover, the scientific validity of emotion recognition itself has been subjected to fundamental critique. \citet{barrett2019expressions} showed in a large-scale review that the inference of emotion from facial movements does not exhibit the one-to-one correspondence that such approaches assume, and critical perspectives on affective AI from data to society have likewise examined the fragility of the foundational premises underlying emotion recognition~\citep{inoshita2026emotion}. These contributions suggest the fragility of the assumption that there is a single true emotion to be read. The present study takes that fragility as a problem of measurement epistemology rather than of normative argument.

\subsection{The Prior Concept of Affective Sovereignty and Its Limitations}

Attempts to thematise the interpretive authority over emotion in the age of emotion AI have appeared only very recently, and the term affective sovereignty itself is of recent coinage rather than an established category. A related study named the principle that the final interpreter of one's emotion is oneself ``affective sovereignty'' and set out formal and computational foundations for implementing that principle within emotion AI systems~\citep{kim2026affective}. That work is wide-ranging. It frames affective sovereignty not as a new fundamental right but as a socio-technical design right that operationalises long-standing commitments to autonomy, dignity, and informational self-determination; it draws on constructed-emotion and appraisal theory to argue that the subject's self-report deserves prima facie authority because an external label is only an imperfect proxy for lived affect; and it builds a computational governance framework, decomposing risk to price interpretive override, proposing a runtime handoff protocol, introducing alignment metrics, and validating the mechanism in proof-of-mechanism simulations. For these reasons it is the prior work that the present study takes most seriously, and engages with directly as its point of departure rather than treating it as one reference among many.

There is, however, one thing that this rich framework does not do, and naming it precisely locates our contribution. The prior work treats the subject's interpretive authority as an ethical commitment: it holds, in its own words, that the person remains the final arbiter ``even when predictive insight is strong.'' What it does not establish is why that authority cannot, even in principle, be discharged by an ever-better measuring system. Its formal apparatus is decision-theoretic, a risk decomposition with an override cost, a manipulation penalty, and an abstention threshold, in which the probability that a user will not endorse the model's label is treated as a primitive quantity; the uncertainty of the meaning itself is never decomposed into a reducible component that more annotation removes and an irreducible component that it cannot. The decisive point lies in this formalism rather than in any implementation limit: even where the prior work notes that its own evaluation setting ``does not capture ambiguity, ambivalence,'' that is offered as a limitation of the simulation, whereas the absence of a reducible--irreducible decomposition is a feature of the formal apparatus itself, the very phenomena an account of irreducibility must address. Consequently the norm is grounded ethically and enforced by design, but it is not derived from the epistemology of measurement; its empirical component is likewise a proof-of-mechanism evaluation of the enforcement architecture, which its author describes as demonstrating feasibility rather than sufficiency, and offers no evidence that individual meaning is in fact unrecoverable, because it does not make that claim.

The present study supplies exactly this missing foundation. Where the prior work assumes the subject's authority and engineers its enforcement, we establish the epistemological proposition that the irreducible component of an individual's meaning distribution cannot be recovered with adequate coverage from finite observation, and we derive interpretive authority from that limit through an explicitly stated non-delegation premise. Our contribution is therefore complementary and occupies a different layer: it is neither the norm, which we adopt, nor its enforcement, which we do not engineer, but the epistemic foundation that explains why the authority cannot be discharged by measurement, however refined. This is also why the present study can replace the ethical answer to the future-prediction worry, that the person remains the arbiter even when prediction is strong, with an epistemic one: for instances whose irreducible component is non-trivial, prediction, however strong, does not recover that component, so there remains nothing for the measuring system to be authoritative about. Where the irreducible component is negligible the measuring system may carry a correspondingly limited authority, and it is precisely the non-trivial residual that the norm protects.

\subsection{Measuring Uncertainty and Ambiguity in Emotion Labels}

Emotion annotation is inherently variable. The phenomenon of multiple annotators assigning different labels to the same stimulus is well known, and evaluation that fixes a single ground-truth label discards this variation as error. \citet{aroyo2015truth} criticised the assumption of a single correct answer as a myth, and \citet{uma2021disagreement} provided a systematic survey of learning from disagreement. \citet{plank2022variation} has argued that variation in human labels should be treated as signal rather than error. That this disagreement can be essential rather than mere noise has been shown through linguistic analysis of disagreement in textual inference~\citep{pavlick2019inherent} and through large-scale collection of collective human judgements on natural language inference~\citep{nie2020chaosnli}. For fine-grained emotion labels, datasets have been developed that record label distributions from many annotators, making the extent of inter-annotator disagreement observable~\citep{demszky2020goemotions}. The report that incorporating human uncertainty as soft labels improves classification robustness~\citep{peterson2019human} further supports the view that the meaning of emotion should be conceived as a distribution rather than a single label.

Part of this variation is a sample-based component that converges as annotator count increases; part is an essential component that persists. To operationally define and estimate the latter, concepts from neighbouring fields have been applied, coverage-based extrapolation from ecology~\citep{chao2012coverage} and the principled lower bound on error in classification~\citep{devroye1996probabilistic}. These approaches share the idea of estimating the limit of infinite draws from finite observations, and can be connected to a framework for measuring the irreducible ambiguity of emotion labels.

The work most closely related to the present study is a line of research that treats the uncertainty of subjective and emotion labels as comprising a component that more data can reduce and a component that it cannot. Decomposing the uncertainty of subjective natural language processing (NLP) into epistemic and aleatoric components via cyclical stochastic-gradient Markov chain Monte Carlo and soft-label learning~\citep{uncertainty_decomp_2026}, showing distributionally that large language models (LLMs) capture the labels of emotion but not its uncertainty~\citep{llm_emotion_uncertainty_2026}, and, in facial expression recognition (FER), separating the aleatoric uncertainty that stems from the intrinsic ambiguity of an expression from the epistemic uncertainty induced by distribution shift and validating that the former tracks annotator disagreement and does not shrink as data grow~\citep{uncertainty_routing_fer_2026}: these studies together provide empirical support for treating the meaning of emotion not as a single label but as a distribution that contains an irreducible component. The present study connects these findings by citation and positions them as the epistemological foundation from which the norm is derived. In doing so, we treat the technical studies that report measurement limits and the normative inquiry into interpretive authority as fulfilling complementary roles.

In summary, existing work has addressed the protection of emotion data, the regulation of emotion recognition, and the declaration and implementation of interpretive authority; but the position of deriving interpretive authority from the epistemological fact that the meaning of emotion cannot in principle be recovered from measurement at the individual-instance level has not been presented. The present study fills this gap and grounds affective sovereignty as a norm anchored in measurement's limits.

\section{The Affectosphere and the Epistemic Gap}\label{sec:gap}

\subsection{Scenario: The Advent of the Affectosphere}

The Affectosphere designates the domain in which emotion is pervasively present, like the atmosphere of society, and becomes both observable and computable. This can be understood as the emotion-focused counterpart of the infosphere, the domain in which information is ubiquitous and reshapes human reality~\citep{floridi2014fourth}. The metaphor is also continuous with discussions of emotion in emotional geography that conceive of feeling as an atmospheric quality of place~\citep{anderson2009atmospheres}; the present study adds to this the new condition of computability. The Affectosphere is characterised not only by the social fact that emotion propagates and shapes relationships, but also by the technological fact that emotion is daily measured, labelled, and made to circulate by AI. The social propagation of emotion has long been noted~\citep{kramer2014contagion}; what is new about the Affectosphere is the entry into a phase in which that propagation is measured at scale, automatically.

There are two stages of AI ubiquity. The first is an active extension of dialogue, in which systems understand what users say. The second is a stage in which the environment passively infers emotion even when users do not explicitly express it. The latter is simultaneously the apex of convenience and the condition in which emotion is continuously read without the user's awareness. As technologies for measuring emotion penetrate vehicles, home appliances, dialogue agents, and urban infrastructure, the emotion labels produced by measurement circulate before the individual's own self-understanding, and can sometimes overwrite it.

This scenario connects to the asymmetries of power that social theory has described in accounts of the commodification of emotion and emotional labour~\citep{zuboff2019surveillance,hochschild1983managed}. In a domain where emotion is measured and commodified, the subject that defines the meaning of emotion quietly shifts from the experiencer to the platform. The question that should be asked within the Affectosphere is not a binary choice of whether to let AI measure emotion at all. Given that measurement is already ubiquitous, the question is: who ultimately determines the meaning of the measured emotion, that is, the locus of interpretive authority.

\subsection{Problem Definition: The Locus of Interpretive Authority}

We formalise the central problem addressed in this study as follows. Consider a situation in which emotion AI outputs an emotion label with accompanying confidence for a given stimulus. The question arises: can the output label be regarded as having determined the meaning of the subject's emotion? We reframe this question as a problem of measurement epistemology. Specifically, we operationally define the meaning of emotion as the distribution of labels that would be assigned by annotators drawn at random from a population under a fixed annotation protocol. We call this distribution the meaning distribution. The meaning distribution is not a single ground-truth label but a probability distribution relativised to the protocol.

Once the meaning distribution is taken as the object of study, the problem of interpretive authority can be specified as a relationship between two quantities. The first is the confidence output by emotion AI, which corresponds to the device's ability to discriminate real differences at the aggregate level of a population. The second is the recoverability of the meaning distribution of an individual instance, to what extent the distribution can be recovered from finite observations. What matters is that there is no guarantee that these two quantities coincide. Whether a device exhibits high confidence and whether the meaning distribution of an individual instance can be determined from finite observations are logically independent claims.

We enter an ontological reservation here. The meaning distribution is an object relativised to a protocol; the present study does not assume a single true value inherent to emotion. As constructivist emotion theory demonstrates, emotion is actively constructed from context, personal history, and conceptual categories, making naively realist claims about a true emotion that can be read unambiguously from outside difficult to sustain~\citep{barrett2017constructed}. Accordingly, we define the subject's authority not as privileged access to a true emotion, but as the procedural final authority to integrate context and give meaning to one's emotion. This procedural definition is formalised in Section~\ref{sec:formal}.

\subsection{Overview: From the Epistemic Gap to the Norm}

\begin{figure}[tbp]
\centering
\includegraphics[width=\textwidth]{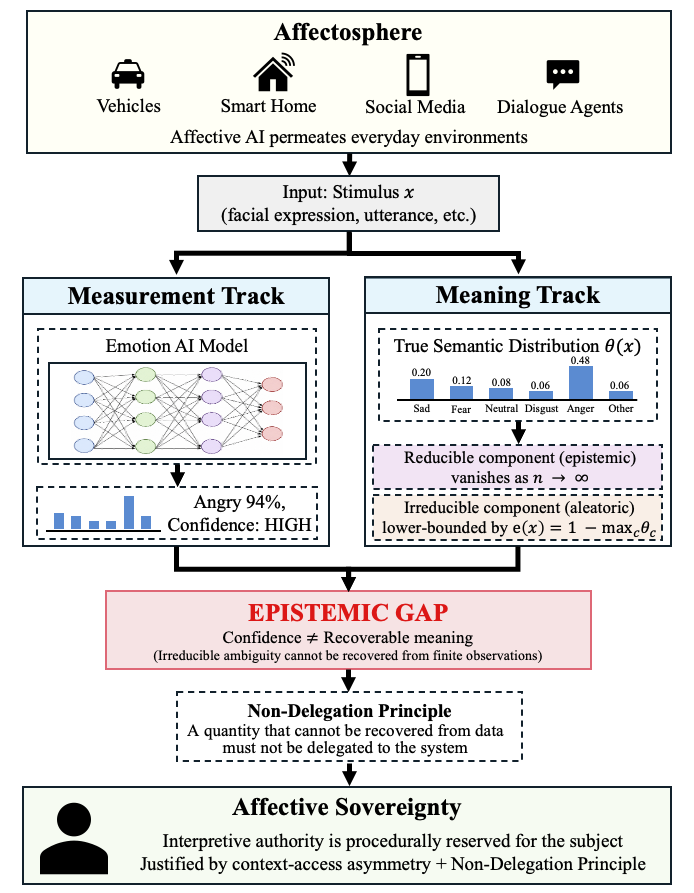}
\caption{The line of argument from the Affectosphere through the epistemic gap to affective sovereignty.}
\label{fig:overview}
\end{figure}

Figure~\ref{fig:overview} illustrates the full line of argument of the present study, from the Affectosphere through the epistemic gap to affective sovereignty. The argument proceeds in two stages. The first stage is an epistemological proposition: measurement systematically diverges from the ability to determine the meaning of individual instances. Emotion AI can assign high-confidence point labels to each instance and discriminate real model differences at the aggregate level. However, the irreducible component of the meaning distribution of an individual instance cannot be estimated with adequate coverage under realistic annotator counts. We call this divergence the epistemic gap.

The second stage is a normative consequence. The irreducible component is not something that can in principle be eliminated by more computation or larger models. Here we explicitly state a normative premise: the output of a system that cannot in principle recover a quantity must not be treated as the authoritative determination of that quantity. Connecting this premise with the epistemic gap yields the conclusion that treating the output of emotion AI as the authoritative determination of irreducible meaning is impermissible. Furthermore, from the asymmetry of access to context, it follows that the remaining interpretive authority belongs procedurally to the experiencing subject. We call this norm affective sovereignty. A distinguishing feature of this study is that the transition from fact to ought is not made without mediation; the normative premise is made explicit before the conclusion is drawn. A formal treatment is given in Section~\ref{sec:formal}.

The remainder of the paper proceeds as follows. Section~\ref{sec:formal} formally defines the meaning distribution and irreducibility, states the epistemic gap as a proposition, and introduces the concept of procedural authority. Section~\ref{sec:demo} empirically grounds this epistemic gap by connecting empirical results from a related study and presenting an analysis showing the overlap between emotion AI confidence and the irreducible domain.

\section{Formalising Irreducibility}\label{sec:formal}

\subsection{Meaning Distribution and Irreducible Ambiguity}

Let $x$ denote an instance representing a stimulus, $C$ the label set, and $K = |C|$ the number of labels. A fixed annotation protocol $\Pi$ includes the label set, instructions, and presentation context. Crucially, $\Pi$ fixes the presentation context and the time of evaluation. We define the distribution of labels assigned to $x$ by annotators drawn at random from a population $\mathcal{P}$ as
\begin{equation}
\theta(x \mid \Pi)
  = \bigl(\theta_1, \dots, \theta_K\bigr),
  \qquad
  \theta_c
  = \Pr_{a \sim \mathcal{P}}\!\left[\,
    a \text{ assigns label } c \text{ to } x
    \;\middle|\; \Pi
  \,\right]
\end{equation}
and call this the meaning distribution. $\theta(x \mid \Pi)$ is not a single ground-truth label; it is an operationalisation of the meaning of emotion relativised to the protocol. In what follows we assume the context is fixed and abbreviate this as $\theta(x)$.

Let $\hat{\theta}_n(x)$ denote the empirical label-proportion vector from $n$ annotators. When each annotator is drawn independently from $\mathcal{P}$, $\hat{\theta}_n(x)$ is an unbiased estimator of $\theta(x)$, and under the fixed protocol
\begin{equation}
  \theta(x) = \lim_{n \to \infty} \hat{\theta}_n(x).
\end{equation}
This limit concerns the draw of annotators under a fixed presentation context and time of evaluation; it does not assert the stability of emotion across contexts. When the ambiguity of the meaning distribution is measured by a diversity functional $D[\cdot]$, we define the irreducible ambiguity as the limiting value $D[\theta(x)]$.

\subsection{Estimation Bias Under Finite Samples and the Reducible Component}

\begin{figure}[tbp]
\centering
\includegraphics[width=0.95\textwidth]{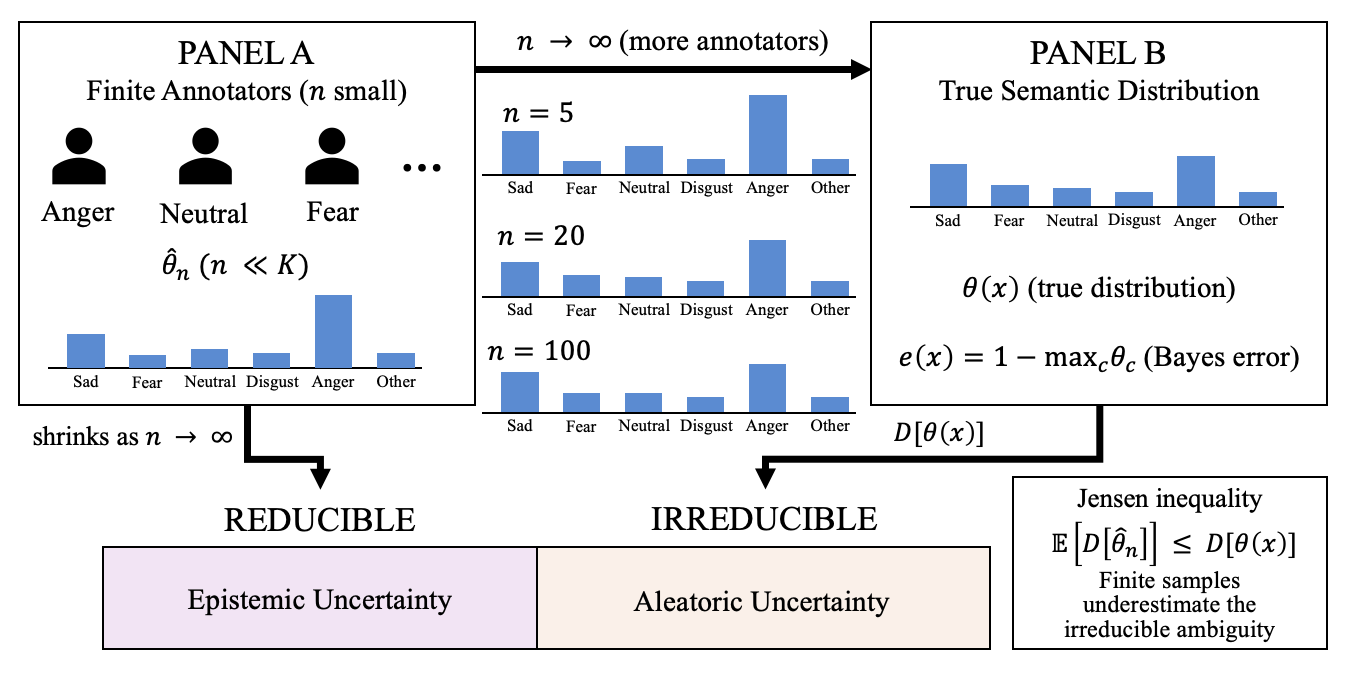}
\caption{Decomposition of the meaning distribution into a reducible (epistemic) component and an irreducible (aleatoric) component.}
\label{fig:decomposition}
\end{figure}

Figure~\ref{fig:decomposition} illustrates the relationship between the observed distribution based on a finite sample and the true meaning distribution, together with the decomposition into reducible and irreducible components. The distinction between reducible and irreducible ambiguity is formalised in terms of estimation convergence. The diversity functionals $D[\cdot]$ used in this study are members of the family of functions that are concave with respect to probability vectors. Specifically, we employ Shannon entropy $H(\theta) = -\sum_c \theta_c \log \theta_c$ and the Gini--Simpson index $1 - \sum_c \theta_c^2$. Both are concave functions of the probability vector $\theta$. (Note: Hill numbers are used interpretively as monotone reparameterisations of these measures; their exponential and reciprocal forms do not in general preserve concavity, so the inequalities below are stated with respect to concave base measures.) From the fact that $\hat{\theta}_n(x)$ is an unbiased estimator of $\theta(x)$ and the concavity of $D$, Jensen's inequality gives
\begin{equation}
  \mathbb{E}\!\left[D\!\left[\hat{\theta}_n(x)\right]\right]
  \;\le\;
  D\!\left[\theta(x)\right],
  \label{eq:jensen}
\end{equation}
with equality achieved in the limit $n \to \infty$. That is, diversity estimates based on finite samples systematically underestimate the irreducible ambiguity $D[\theta(x)]$. This direction of underestimation is consistent with related work that decomposes the uncertainty of subjective labels into a reducible and an irreducible component~\citep{uncertainty_decomp_2026}.

We define the reducible component as the estimation bias and variance that vanish as annotator count $n$ increases. Specifically, the estimation bias $\Delta_n(x) = D[\theta(x)] - \mathbb{E}[D[\hat{\theta}_n(x)]] \ge 0$ converges to zero as $n \to \infty$. In contrast, the irreducible component $D[\theta(x)]$ persists no matter how many annotators are added. Coverage-based extrapolation~\citep{chao2012coverage} provides one approach to estimating $D[\theta(x)]$ from finite observations; however, as we discuss below, it is known that this extrapolation is not reliable for individual instances in fine-grained, small-sample settings. A non-trivial lower bound on the irreducible component is given by the error of the optimal predictor, namely $e(x) = 1 - \max_c \theta_c(x)$~\citep{devroye1996probabilistic}. For instances where $e(x) > 0$, the meaning cannot in principle be collapsed to a single label. This decomposition is the methodological pivot that distinguishes the present account. Prior governance-oriented work treats the probability that a user disagrees with the model's label as a primitive quantity to be priced as a cost~\citep{kim2026affective}; here, by contrast, we open that quantity up, separating the part that more annotation removes from the part $D[\theta(x)]$ that it cannot, and it is the latter that grounds interpretive authority.

This distinction corresponds to the taxonomy of uncertainty in machine learning. The reducible component corresponds to epistemic uncertainty, which can be reduced with more data; the irreducible component corresponds to aleatoric uncertainty, which cannot be reduced by adding data~\citep{hullermeier2021aleatoric}. The correspondence must be read relative to the protocol. Some treatments classify annotator disagreement as epistemic, on the ground that it reflects the modeller's limited knowledge of individual annotators; here, however, the relevant quantity is the disagreement that remains under a fixed protocol as the number of annotators grows without bound. That residual is irreducible by further annotation under the same protocol, and it is this protocol-relative residual that we identify as aleatoric, while the estimation uncertainty about $\theta(x)$ that more annotators do remove is the epistemic part. The present study's claim can then be organised as a twofold assertion: that the meaning distribution of emotion contains aleatoric uncertainty at a non-trivial level, and that estimating it for individual instances is itself difficult.

We illustrate with a concrete example. Suppose that five annotators assign emotion labels to a text stimulus, with the distribution: ``confused'' (2 votes), ``disappointed'' (2 votes), and ``neutral'' (1 vote). The empirical distribution is $(0.4, 0.4, 0.2)$. In a fine-grained label set with more than twenty categories, the large majority of unobserved labels receive zero, and from these five votes it is nearly impossible to infer the shape of the true meaning distribution $\theta(x)$. We distinguish two cases. If $\theta(x)$ is in fact balanced among several emotions, that balance is irreducible: no matter how many annotators are added, the meaning distribution will not converge to a single label. If, on the other hand, the distribution is skewed toward one emotion but the votes happened to be split, then even determining which of these cases applies is impossible from five observations alone. In either case, when the device classifies the stimulus as ``confused'' with high confidence, that is merely an output that collapses the meaning distribution of the individual instance to a point; it does not imply that the irreducible component of $\theta(x)$ has been recovered. In the former case the irreducible component cannot in principle disappear; in the latter it cannot be identified from finite observations. This example shows that the ontological route, by which the irreducible component cannot in principle vanish, and the epistemological route, by which the irreducible component cannot even be identified at the individual level, each independently constitute the epistemic gap.

\subsection{The Epistemic Gap}

We distinguish two quantities. The first is device confidence: the point label $\hat{y}(x)$ that emotion AI assigns to each $x$, together with the accompanying score that appears to be well-calibrated. At the aggregate level of the population, this output exhibits high ability to discriminate real differences. The second is recoverable meaning: to what extent the irreducible ambiguity $D[\theta(x)]$ of an individual instance $x$ can be estimated with adequate coverage from finite observations.

\begin{proposition}[Epistemic Gap]
Device confidence and the recoverability of the meaning of an individual instance systematically diverge. That is, no matter how refined measurement becomes, the set of instances for which the irreducible ambiguity $D[\theta(x)]$ cannot be determined with adequate coverage from finite observations exists at a non-trivial proportion. Consequently, high confidence does not constitute evidence that an irrecoverable quantity has been recovered.
\end{proposition}

The proposition rests on two pillars of different logical status, and keeping them apart is essential. The first is a statistical fact that holds independently of any dataset: for instances where $e(x) > 0$, the irreducible component $D[\theta(x)]$ is strictly positive, and, as Equation~\eqref{eq:jensen} shows, any finite-sample plug-in systematically underestimates it. This pillar establishes that the irreducible component exists and is not captured by the point estimate; it is a theorem, not an empirical conjecture. The second pillar concerns recoverability and is structural rather than purely deductive. Recovering $D[\theta(x)]$ at the level of an individual instance requires estimating a $K$-dimensional distribution from $n$ observations; when $n \ll K$ the empirical distribution $\hat{\theta}_n(x)$ is supported on at most $n$ of the $K$ categories, so the functional is severely under-constrained and its estimator carries large bias and variance. We therefore do not claim that a valid confidence interval for $D[\theta(x)]$ is achievable per instance; on the contrary, individual-level recovery with nominal coverage is, in general, not achievable in this regime, a consequence of the ill-posedness rather than of any particular estimator. The related study cited in Section~\ref{sec:demo} corroborates the severity of this failure in realistic settings, across multiple functionals and confidence interval methods. This division of labour also marks an asymmetry with engineering-oriented accounts: because the existence of the irreducible component is a theorem, our empirical material illustrates the second pillar rather than being the sole support for the claim, whereas a proof-of-mechanism evaluation can demonstrate feasibility but not such an in-principle limit.

\subsection{Procedural Authority as Interpretive Authority}

In deriving normative consequences from the epistemic gap, we take seriously the danger of passing from fact to ought without mediation~\citep{hume1739treatise} and explicitly state one normative premise.

\begin{quote}
Principle of Interpretive Non-Delegation. For a quantity $Q$: the output of a system that cannot in principle recover $Q$ must not be treated as the authoritative determination of $Q$.
\end{quote}

This premise can be defended as an extension of two established norms. The first is an epistemic norm: one must not grant authority to readings beyond the resolution of a measuring instrument, that is, one must not place undue epistemic reliance on outputs that exceed the system's capacity. The second is a norm of self-determination: the authority to decide matters that intimately concern oneself should be attributed to the individual herself, as systematised in the right to informational self-determination~\citep{westin1967privacy} and the principle of respect for autonomy in medical ethics~\citep{beauchamp2019principles}. Since the determination of irrecoverable emotional meaning both intimately concerns the individual and is something that no external system can in principle perform, both norms jointly exclude delegating that determination to the system. The Epistemic Gap Proposition satisfies the condition of this premise, namely, that the system cannot recover the irreducible component, and therefore treating the output of emotion AI as the authoritative determination of irreducible meaning is impermissible.

We next argue, on two grounds, to whom the authority over this irrecoverable meaning-making belongs. The positive ground is the asymmetry of access. From the perspective of procedural authority, irreducible meaning is conferred by the integration of context, personal history, and goals; among those available, the only one with privileged access to this information is the experiencing subject. The negative ground is harm minimisation. Given that erroneous external attribution can deform the individual's self-understanding~\citep{hacking2007kinds}, placing interpretive authority as a default in a fallible external system generates the harm of eroding the subject's self-understanding. The default placement that minimises harm is therefore to reserve interpretive authority with the subject. By these two grounds, the final interpretive authority that remains under the Principle of Interpretive Non-Delegation belongs, as a default, to the experiencing subject.

Here the subject's authority is understood not as infallible access to a true emotion, but as the procedural final authority to integrate context and give meaning to one's own emotion. This understanding is aligned with the philosophical view that first-person authority in self-knowledge is not infallible observation through introspection, but rather the stance of a subject who takes responsibility for their own attitudes~\citep{moran2001authority}. The procedural authority in this study is similarly the authority of a subject who bears responsibility for contextual integration, not of a privileged observer. This definition is consistent with the constructivist emotion theory's finding that no single true emotion stable across contexts exists~\citep{barrett2017constructed}: defining $\theta(x)$ as the sample limit under a fixed context does not contradict the constructivist claim that emotion is reconstituted across contexts. The two operate at different levels, within-context annotator variation and between-context variation, respectively. Furthermore, appraisal theory's finding that emotion depends on the subject's appraisal~\citep{lazarus1991emotion,scherer2009dynamic,moors2013appraisal}, and the observation that erroneous external attribution of emotion can deform the subject's self-understanding~\citep{hacking2007kinds}, both support the location of procedural authority. This procedurally defined interpretive authority is what has been called affective sovereignty~\citep{kim2026affective}; what the present study adds is its derivation from the limits of measurement rather than its assertion as a right.

\section{Empirical Grounding}\label{sec:demo}

Although the present study is centred on conceptual analysis, its claims are grounded in empirical findings. In this section, we demonstrate that the epistemic gap is not a theoretical construction but holds in real emotion-label data, drawing on a body of empirical work on the uncertainty of emotion labels.

\subsection{The Inestimability of the Irreducible Component}

The first foundation of the epistemic gap is that the irreducible component of individual instances cannot be estimated under realistic annotator counts. As argued in Section~\ref{sec:formal}, the logical pillar of this claim rests on statistical facts that do not depend on any specific dataset; but the severity of this failure in realistic settings can be empirically confirmed. The validity of the proposition itself does not depend on these results, as noted above; but they show empirically that the proposition is not a mere construction.

First, LLMs reproduce the labels of emotion with high accuracy while failing to capture the underlying distribution of uncertainty, as shown by a distributional analysis of human--model judgment gaps~\citep{llm_emotion_uncertainty_2026}; this directly supports the claim that measurement yields a point label without recovering the distribution of meaning. Second, the uncertainty of subjective NLP decomposes into an epistemic component that more data can reduce and an aleatoric component that it cannot~\citep{uncertainty_decomp_2026}. Third, work that decomposes the uncertainty of FER into aleatoric and epistemic components shows, by validation against human disagreement, that the aleatoric component tracks annotator disagreement and does not shrink as data grow, and at the same time that a single uncertainty score conflates the two~\citep{uncertainty_routing_fer_2026}. That conflation corresponds precisely to the structure of the epistemic gap, in which device confidence is mistaken for recoverable meaning. Together, these findings empirically substantiate the distinction between a component reducible by aggregation or further data and an irreducible component that remains for individual instances, in keeping with the structure of the epistemic gap.

\subsection{The Relationship Between Confidence and Irreducibility}

The epistemic gap can also be examined from the side of device confidence. What matters here is what confidence measures. To the extent that it is calibrated, the confidence that emotion AI outputs reflects the reliability of a prediction under the training distribution, that is, the smallness of epistemic uncertainty, and not whether the meaning distribution of the instance is concentrated on a single label, that is, the smallness of aleatoric uncertainty. These two are in principle independent. Consequently, however much the calibration of confidence is improved, this only makes confidence an honest expression of predictive reliability; it does not reduce the irreducible aleatoric component.

This independence is consistent with the underestimation that Equation~\eqref{eq:jensen} establishes. Since diversity estimates based on finite samples systematically underestimate the irreducible component, a device that concentrates its confidence on a single label tends, if anything, to make the irreducible component less visible. As a result, it is in principle expected that instances whose meaning distribution is essentially split among several emotions, that is, instances with large irreducible components, will be found among instances to which the device assigns high confidence. The stimulus in the worked example of the previous section, where confusion and disappointment are in contention, is an instance of this type: the device may classify it as confusion with high confidence while its irreducible component is large.

Quantifying this expectation on real data reinforces the claim of this study from the side of confidence. Concretely, for an emotion dataset with a distribution of labels assigned by many annotators, one would juxtapose the confidence output by emotion AI with an estimate of the irreducible component of each instance, and measure the proportion of instances that are both high-confidence and highly irreducible. A non-trivial mass of instances in this region would directly show the epistemic gap, namely that high confidence does not entail the recovery of meaning. This quantification and visualisation are not part of the core argument of this study, but constitute a natural next step that would reinforce it empirically, and are identified as a direction for future work.

In summary, the core of the epistemic gap, the claim that the irreducible component of individual instances cannot be recovered from finite observations, is supported by the empirical results of the related study. The present study builds the normative consequences of Section~\ref{sec:formal} on this empirical foundation.

\section{Objections and Replies}\label{sec:objections}

The present study's central claim, the argument deriving affective sovereignty from the epistemic gap, is subject to a number of anticipated objections. In this section we address five objections that appear most significant. They fall into two groups. The first group targets the novelty of our contribution and the validity of the argument, and the second targets the scope of our claim and its metaphysical implications. Addressing them in these two groups clarifies the scope and limits of the argument.

\subsection{Objections to the Novelty and Validity of the Argument}

Objection 1: Affective sovereignty is merely a restatement of existing neurorights. Cognitive liberty, mental privacy~\citep{ienca2017newrights}, and the right to mental self-determination~\citep{bublitz2014crimes} have already systematised rights over the mental domain; does the present study add anything? Reply. The two differ in both the object of protection and the direction of argument. Neurorights are primarily directed at protecting the mind from being read or manipulated. Affective sovereignty does not take issue with the reading of emotion as such; it asks who determines the irreducible meaning that remains after the reading, the locus of interpretive authority. Moreover, the contribution of this study is not to demand a right, but to derive it from measurement's limits. Affective sovereignty is therefore positioned not as an alternative to neurorights but as a complementary extension that adds the dimension of interpretive authority.

Objection 2: The Principle of Interpretive Non-Delegation is a question-begging premise introduced to secure the conclusion. Reply. This premise is not specific to emotion; it is a general norm that is independently motivated. Consider a scale with a minimum graduation of one gram: treating the scale's display as the authoritative determination of a difference smaller than one gram would be an error. The error arises not because the scale is defective, but because the device is being asked to adjudicate a matter beyond its resolution. The Principle of Interpretive Non-Delegation holds as a general epistemic norm in this emotion-independent case as well. Crucially, the premise does not imply the conclusion directly; it is applied to the emotion case only after the Epistemic Gap Proposition is independently established in Section~\ref{sec:formal}. The premise does not assume the fact that emotion AI cannot recover the irreducible component; that fact is shown separately. Furthermore, the premise converges with the norm of self-determination, the attribution of decision authority over matters concerning oneself to the individual~\citep{westin1967privacy,beauchamp2019principles} , which provides an independent motivation. Since the premise can be defended independently of the emotion case, and its condition of application is established independently, the charge of question-begging does not hold.

\subsection{Objections to the Scope and Metaphysics of the Claim}

Objection 3: If measurement is valid at the aggregate level, that is sufficient for practical purposes. Reply. Aggregate-level validity and individual-level authority are distinct claims. Many deployments of emotion AI act on individual users, not on an aggregate population. An in-vehicle system that adjusts ambient sound when it judges the driver to be frustrated, or a dialogue agent that changes its responses based on inferred user emotion, each makes decisions about individual instances~\citep{andalibi2020human}. The epistemic gap arises at exactly this individual level, the level at which deployments actually operate, while the level at which measurement can be valid differs from it. Aggregate-level validity therefore does not justify delegating interpretive authority at the individual level.

Objection 4: If constructivist emotion theory is invoked, the subject's own self-understanding is also destabilised, and the subject's authority is equally undermined. Reply. This study defines the subject's authority as procedural authority, not as privileged observation~\citep{moran2001authority}. What constructivism undermines is the image of a privileged observer who reads emotion infallibly; it does not undermine the image of a responsible subject who integrates context and gives meaning to their own emotion. The subject's authority derives not from infallible access to a fixed internal fact, but from privileged access to the context that constitutes meaning and from the standing to bear responsibility for its integration. Constructivism therefore undermines the authority of external measurement while leaving the subject's procedural authority intact.

Objection 5: Does the claim that irreducible meaning cannot be recovered externally entail anti-realism or solipsism about emotion? Does saying that irreducible meaning cannot be externally recovered amount to saying that emotion has no objective reality? Reply. The claim is an epistemological modesty about individual-level recoverability, not a metaphysical claim about the existence of emotion. The meaning distribution $\theta(x)$ is a well-defined object relativised to a protocol, and aggregate regularities are real. Protocol-relativity does not stand in opposition to reality. Just as many scientific quantities are defined relative to measurement conventions while being objectively real, $\theta(x)$ is a real object uniquely determined under the fixed protocol. Being protocol-relative means that the conditions of measurement are explicit; it does not mean the object is arbitrary or non-real. What the present study denies is the assumption that external measurement can determine the irreducible component of an individual instance, not the reality of emotion itself. This distinction is consistent with the general understanding of the concept of uncertainty, according to which acknowledging aleatoric uncertainty does not entail that the object in question does not exist~\citep{hullermeier2021aleatoric}.

\section{Discussion}\label{sec:discussion}

\subsection{Implications for the Design and Evaluation of Emotion AI}

The epistemic gap provides concrete guidance for the design and evaluation of emotion AI. First, the unit and the metric of evaluation need to be reconsidered. Since exhibiting high discrimination performance at the aggregate level of a population is a different matter from determining the meaning of an individual instance, high confidence for individual instances should not be used uncritically as a performance metric. In particular, an evaluation that fixes a single ground-truth label as the true value and measures performance by agreement with it discards the irreducible component as error, and may even reward a model that produces high confidence on essentially split instances. To avoid this, evaluation should also measure agreement with the soft-label distribution and report the reducible and irreducible components separately, disclosing individual-level coverage independently of aggregate accuracy~\citep{uma2021disagreement,plank2022variation}.

Second, the form of the output and the meaning of confidence need to be designed. For instances with large irreducible components, a design in which the device suppresses confidence, presents the meaning distribution itself rather than a single label, or reserves interpretation and defers to the subject is preferable. Confidence should be calibrated as an honest expression of predictive reliability, and must not be used as a quantity that conceals irreducible ambiguity. Third, the output of emotion AI should be positioned not as a determined meaning but as one input that supports the subject's procedural authority. This requires that the boundary between inference and intervention be made explicit by design: a distinction between the device inferring an emotion and intervening in the subject on the basis of that inference, with the subject's interpretation taken as the final reference point for the latter. Existing guidance on the ethics of emotion recognition~\citep{mohammad2022ethics} likewise supports the need for such design caution. These implications also connect the present account to the engineering frameworks built on the same norm. Where such a framework prices the disutility of contradicting the user's self-report as an override cost~\citep{kim2026affective}, our analysis supplies the rationale for why that cost should exist at all and where it should be largest: the override cost is warranted precisely on instances whose irreducible component is large, because there the system has, in principle, no authority to recover. In this sense the epistemic foundation does not merely sit beside the design layer but grounds one of its central design choices, indicating where a system must defer.

\subsection{Implications for Regulation and Institutional Design}

The normative consequences of this study complement existing regulatory frameworks. Protection of emotion data and restrictions on the use of emotion recognition in workplace and educational contexts~\citep{euaiact2024} regulate whether and where emotion AI may be used; yet each concerns the permissibility and location of use, and none answers the question of who ultimately determines the meaning of the measured emotion in contexts where use is permitted. Making affective sovereignty explicit as a design criterion provides an institutional answer to this remaining question.

Concretely, institutionalising affective sovereignty requires that entities deploying emotion AI guarantee procedures by which the subject can ultimately determine the interpretation of their own emotion and can correct or reject the attribution made by the device. This is an extension, to the affective domain, of the way in which rights of rectification and erasure in the informational domain have reserved final control over data concerning oneself to the subject. The procedure further requires that accountability be made explicit for each deployment context: which emotion the device attributed and with what confidence, how that attribution is disclosed to the subject, and how the subject can correct it must be built into the design of the deployment. Such requirements can be positioned as a natural extension of the trajectory by which the right to informational self-determination~\citep{westin1967privacy} and respect for autonomy in medicine~\citep{beauchamp2019principles} have been institutionalised, and they outline a new form of procedural safeguard for an era in which emotion-processing technologies become social infrastructure.

\subsection{Implications for the Epistemic Status of Emotion AI}

Beyond design and regulation, the epistemic gap bears on how we should regard emotion AI as a knowledge-producing system, a question of the epistemic status of a computational discipline rather than of its engineering. The gap implies that the outputs of emotion AI have a dual epistemic status. At the aggregate level, where it discriminates real differences across a population, emotion AI is a legitimate instrument that produces reliable knowledge. At the level of the individual instance, however, where the irreducible component is non-trivial a confident point label does not have the epistemic status of a determination of meaning; it is a prediction of the most probable label under the annotation protocol, whose irreducible component remains unresolved. To treat the former kind of output as if it were the latter is not merely a practical mistake but a conflation of two distinct epistemic statuses of the output, a category error in the epistemological rather than the Rylean ontological sense.

This reframing follows from importing the distinction between epistemic and aleatoric uncertainty into the affective domain~\citep{hullermeier2021aleatoric}. That distinction reclassifies the kind of knowledge emotion AI can and cannot produce: reducing epistemic uncertainty, through more data or better models, sharpens what the system can legitimately claim, whereas the aleatoric, irreducible component marks a boundary that no amount of computation crosses. The originality of the present study lies in carrying this technical distinction through to a normative consequence: once the individual-level output is seen as a bounded prediction rather than a determination, the authority to fix the meaning cannot rest with the instrument. Emotion AI, on this view, is best represented not as an oracle of affective meaning but as a probabilistic instrument whose authority is bounded by what it can recover, and whose proper role is to inform, not to displace, the subject's own interpretation.

\subsection{Limitations}

The present study has several limitations. First, the meaning distribution is an object relativised to a protocol and depends on the design of that protocol. The size of the irreducible component can change under different instructions or presentation contexts, and the claims of this study are therefore claims under the condition of a fixed protocol. Second, the normative consequences of this study depend on the Principle of Interpretive Non-Delegation. Although we have defended this premise as an instance of the general epistemic norm that authority must not be granted to readings beyond an instrument's resolution, the normative strength and scope of the premise itself remain open to further discussion. Similarly, the step by which the remaining interpretive authority is attributed to the experiencing subject depends on the premise of asymmetric access to context; the refinement of attribution in situations where multiple parties share context is a direction for future work. Third, the analysis that quantifies the epistemic gap from the side of device confidence, directly visualising the relationship between emotion AI confidence and the irreducible component on real data, is only conceptually positioned in this study; its empirical completion is also identified as a direction for future work.

\section{Conclusion}\label{sec:conclusion}

The present study addressed the problem of the locus of interpretive authority, who ultimately determines the meaning of emotion, within the Affectosphere, the domain in which emotion-sensing AI is becoming pervasively embedded in social infrastructure. In response to this question, we approached it from the epistemological side of measurement's structural limits. We defined a meaning distribution under a fixed annotation protocol, provided a framework that decomposes the uncertainty of this distribution into reducible and irreducible components, and formalised as a proposition the systematic divergence between device confidence and the recoverability of individual-instance meaning, the epistemic gap.

As empirical grounding, we connected by citation the finding of a related study, that the irreducible component of individual instances cannot be estimated with adequate coverage under realistic annotator counts, and demonstrated the structure in which aggregate-level measurement is possible while individual meaning is not. By joining this epistemic gap with the normative premise that the output of a system which cannot in principle recover a quantity must not be treated as its authoritative determination, we derived the norm that the irreducible meaning of emotion cannot be delegated to a measuring device and is procedurally reserved, by virtue of the asymmetry of access to context, for the experiencing subject, the norm of affective sovereignty. Because the transition from fact to ought was not made without mediation but with the premise made explicit, this derivation demonstrates that the design, evaluation, and regulation of emotion AI should place the explicit allocation of interpretive authority, rather than the maximisation of accuracy, at their core.

Directions for future work are threefold. First, we will complete the analysis showing the overlap between emotion AI confidence and the irreducible component, thereby reinforcing the empirical demonstration of the epistemic gap. Second, we will philosophically scrutinise the normative strength of the concept of procedural authority, clarifying the reach of the deduction from irrecoverability to non-delegability. Third, we will examine the behaviour of the irreducible component under different protocols and emotion models, and construct concrete procedures for operationalising affective sovereignty as a design criterion.